\documentclass[runningheads]{llncs}
\usepackage[T1]{fontenc}
\usepackage{xcolor}
\usepackage{graphicx}
\usepackage{amsmath}
\usepackage{enumitem}
\usepackage{tabularx}
\usepackage{pifont}
\usepackage{booktabs}
\usepackage{amssymb}
\usepackage{subfig}
\begin{document}
\title{PaliGemma-CXR: A Multi-task Multimodal Model for TB Chest X-ray Interpretation}
\titlerunning{PaliGemma-CXR}
%
\author{Denis Musinguzi\inst{1} \and Sudi Murindanyi\inst{1} \and
Andrew Katumba\inst{1}
}
\authorrunning{Denis Musinguzi and Andrew Katumba}
%
\institute{Makerere University\\ 
\email{musinguzidenis97@gmail.com, murindanyi@gmail.com, andrew.katumba@mak.ac.ug}
}

\maketitle              
\begin{abstract}

Tuberculosis (TB) is a infectious global health challenge. Chest X-rays are a standard method for TB screening, yet many countries face a critical shortage of radiologists capable of interpreting these images. Machine learning offers an alternative, as it can automate tasks such as disease diagnosis, and report generation. However, traditional approaches rely on task-specific models, which cannot utilize the interdependence between tasks. Building a multi-task model capable of performing multiple tasks poses additional challenges such as scarcity of multimodal data, dataset imbalance, and negative transfer. To address these challenges, we propose PaliGemma-CXR, a multi-task multimodal model capable of performing TB diagnosis, object detection, segmentation, report generation, and VQA. Starting with a dataset of chest X-ray images annotated with TB diagnosis labels and segmentation masks, we curated a multimodal dataset to support additional tasks. By finetuning PaliGemma on this dataset and sampling data using ratios of the inverse of the size of task datasets,  we achieved the following results across all tasks: 90.32\% accuracy on TB diagnosis and 98.95\% on close-ended VQA, 41.3 BLEU score on report generation, and a mAP of 19.4 and 16.0 on object detection and segmentation, respectively. These results demonstrate that PaliGemma-CXR effectively leverages the interdependence between multiple image interpretation tasks to enhance performance. 

\end{abstract}
\section{Introduction}

Tuberculosis is one of the most infectious diseases worldwide and causes millions of deaths~\cite{Bloom2017}. Prior to the COVID-19 pandemic, it was the leading infectious cause of mortality worldwide~\cite{Bloom2017}. Chest X-rays play a critical role in the detection and diagnosis of TB, particularly because of low cost and speed. However, the traditional approach of relying on expert clinicians, typically radiologists, to interpret these images is both costly and impractical, especially in low-income countries where there is a severe shortage of skilled professionals. Deep learning  has emerged as a promising alternative, achieving remarkable success in various medical imaging tasks. Some of these models match or even exceed the performance of human experts.

Most traditional approaches tackle medical image interpretation tasks in isolation, resulting in separate models with limited generalization ability and efficiency. 
However, the performance on individual tasks can be improved by jointly training a single model on multiple tasks. 
This performance boost arises from the interdependence between tasks, where features learned from one task can enhance the model's ability to perform another~\cite{Caruana1997}. 
However, achieving the performance improvement can be challenging because of scarcity of multimodal datasets covering multiple tasks, imbalance in the size of the task-specific datasets, and negative transfer between the tasks~\cite{Standley2019}.

In this work, we propose PaliGemma-CXR, a multi-task multimodal model that can perform multiple medical image interpretation tasks such as disease diagnosis, object detection, segmentation, visual question answering, and report generation. 
We solve the data problem by leveraging labeled data from a clinical study on TB patients in Uganda. Starting with a dataset labeled with TB diagnosis and segmentation masks of features in the X-ray images, we generate a dataset containing bounding boxes, medical reports and questions and answers about the images.
We address dataset imbalance by sampling using the ratio of the inverse of the task-specific datasets~\cite{perera2018}.
We finetune PaliGemma on all the tasks in the derived datasets jointly.
PaliGemma-CXR outperformed the task-specific model across all tasks. For
disease diagnosis and object detection, the model outperformed vision only
models.

In summary, our contributions are as follows:
\begin{enumerate}
    \item We curate a multimodal dataset covering medical disease diagnosis, visual question answering, radiology report generation, object detection and image segmentation tasks. 

    \item We introduce PaliGemma-CXR, a single multi-task multimodal model that can perform medical disease diagnosis, visual question answering, radiology report generation, object detection and image segmentation with the same set of model weights.

    \item We demonstrate that a single multi-task multimodal model outperforms task-specific models on all medical interpretation tasks.
\end{enumerate}

\section{Related Work}

The application of deep learning to tuberculosis (TB) medical imaging has gained significant traction, with models often surpassing radiologist performance in specific tasks. However, the majority of research in this domain has focused on disease diagnosis, determining whether an image shows signs of TB or not. These studies predominantly rely on a limited number of datasets, such as the Shenzhen and Montgomery datasets~\cite{Jaeger2014}, which, while valuable, do not fully represent the diversity of TB cases, particularly in high-burden regions like Africa. For instance, Dasanayaka et al.~\cite{Dasanayaka2021} and Acharya et al.~\cite{Acharya2022} leveraged these datasets to build TB diagnosis models.
A few studies like Kawuma et al.~\cite{Kawuma2024} and Rajpurkar et al.~\cite{Rajpurkar2020} utilized custom datasets collected from Uganda and South Africa, respectively, demonstrating the potential of deep learning in real-world, high-burden settings. Despite these advances, the scarcity of locally collected data remains a significant challenge, limiting the generalization and applicability of these models in regions where TB is most prevalent.

The emergence of powerful generalist vision-language models~\cite{radford2021learningtransferablevisualmodels,liu2023visualinstructiontuning,alayrac2022flamingovisuallanguagemodel} has expanded the capabilities of machine learning by unifying vision and language tasks. These models leverage the flexibility of language to specify a wide range of tasks using a unified output space [17]. This approach enables traditionally distinct tasks, such as object detection and segmentation, to be addressed jointly through language-based outputs [4]. This paradigm has sparked interest in adapting these models to the medical domain~\cite{moor2023medflamingomultimodalmedicalfewshot,li2023llavamedtraininglargelanguageandvision,zhang2024pmcvqavisualinstructiontuning}, which are fine-tuned on large datasets of paired medical image-text data extracted from publications and textbooks.

Several multi-task and multimodal biomedical models have been developed to handle diverse tasks in the medical domain. For instance, BiomedGPT~\cite{Zhang_2024}  and ELIXR~\cite{xu2023elixrgeneralpurposexray}, utilize language-aligned image encoders to perform multiple image interpretation tasks. However, these models lack the ability to handle both text-based and vision-based tasks seamlessly within a unified framework. They either perform vision only or language only tasks. BiomedParse~\cite{zhao2024biomedparsebiomedicalfoundationmodel} performs multiple medical interpretation tasks across multiple modalities but is designed for task-specific outputs.

Among existing models, Med-PaLM M~\cite{tu2023generalistbiomedicalai} is the most similar to our work.
However, it is primarily limited to text-based tasks, whereas our approach extends this capability to both text-based and vision-based tasks, enabling a more comprehensive and flexible solution for medical image analysis.
\section{Dataset}
\subsection{Data Collection}
\label{dataset}
Lack of open chest X-ray datasets from Africa has hindered the development of deep learning models tailored for the local context. To address the problem, we conducted a clinical study to collect chest X-ray images from suspected TB patients in rural and urban areas in Uganda. We collected digital chest X-rays from Mengo and Mulago hospitals in urban areas, and Amai and Kisiizi hospitals in rural areas for over a year. The study was approved by the Mengo Hospital Research and Ethics Committee prior to data collection, including consent and data sharing agreements. Additional approval was sought from the Uganda National Council for Science and Technology before commencing with data collection. Every images in the dataset was reviewed by 2 trained radiographers and medical officers. The entire dataset was composed of 1,835 images from the four locations. The dataset includes 902 images from Ernest Cook Ultrasound Research and Education Institute, 319 images from Kisiizi hospital, 716 images from Mulago hospital and 91 images from Mengo hospital.

\subsection{Data Annotation}
Three radiologists with at least 10 years experience to annotate the
images. We split the images among the radiologist with each image being labeled
by 2 radiologists. The radiologists classified the quality of the images as good,
average and poor. The radiologists then annotated segmentation masks around the
pathologies observed in the image. The radiologists categorized the chest X-ray
images into 3 classes related to TB infection: TB positive, TB negative
and sick but no TB. They further classified the TB positive images
as either active TB or inactive TB. In the final dataset, only images where the two radiologists were in agreement and of good or average quality were included. 
The resulting dataset had 1,149 images categorized as follows: 715 under active TB, 69 under inactive TB, 332 under normal and 33 under sick but no TB.

\subsection{Multimodal Datasets}
Starting with the data described above, we created datasets for report generation, visual question answering and segmentation. 

\begin{figure}[!htb]
\minipage{0.2\textwidth}
  \subfloat[]{\label{main:a}\includegraphics[scale=.075]{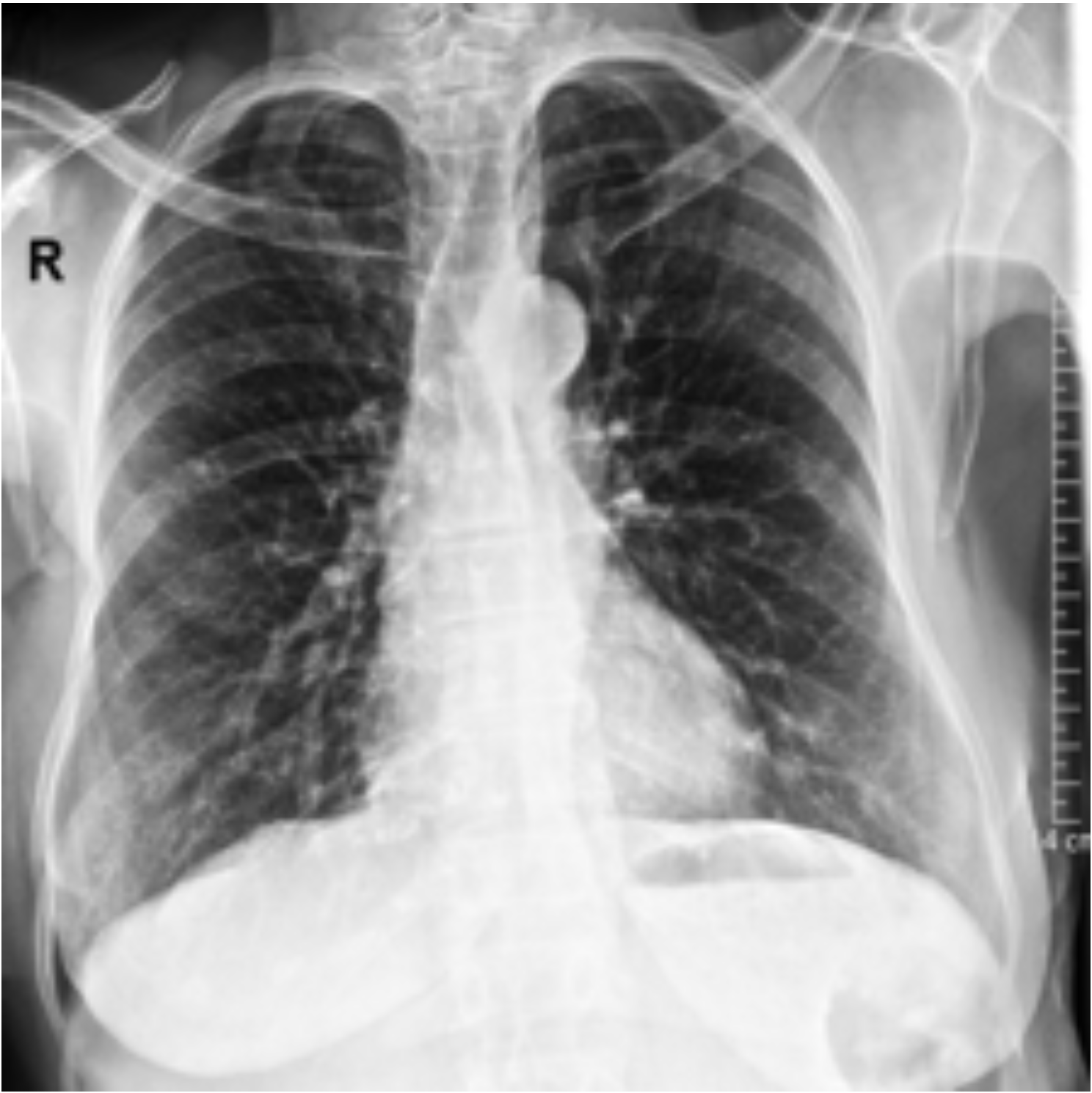}}
\endminipage\hfill
\minipage{0.2\textwidth}
  \subfloat[]{\label{main:a}\includegraphics[scale=.075]{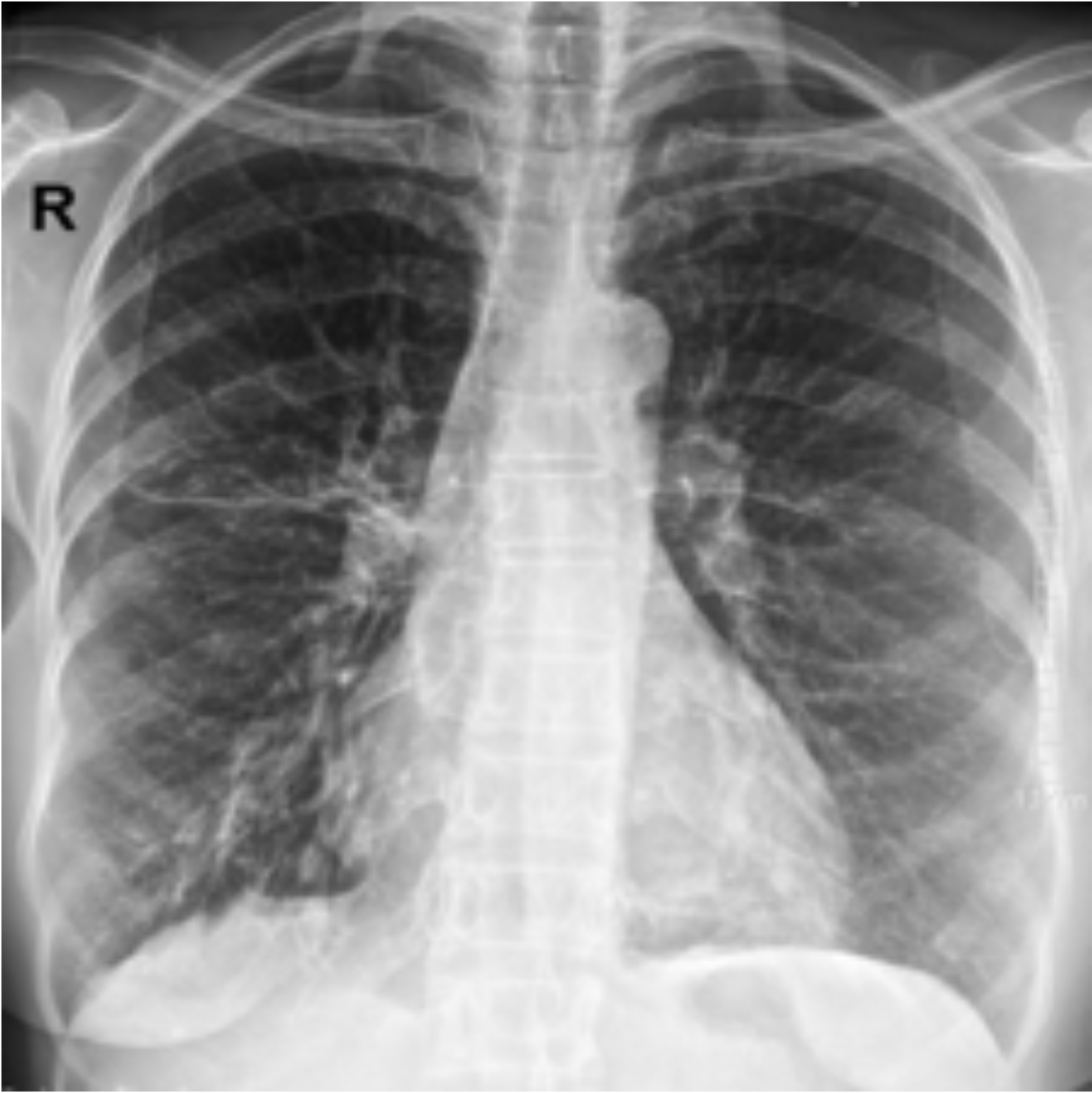}}
\endminipage\hfill
\minipage{0.2\textwidth}%
  \subfloat[]{\label{main:a}\includegraphics[scale=.075]{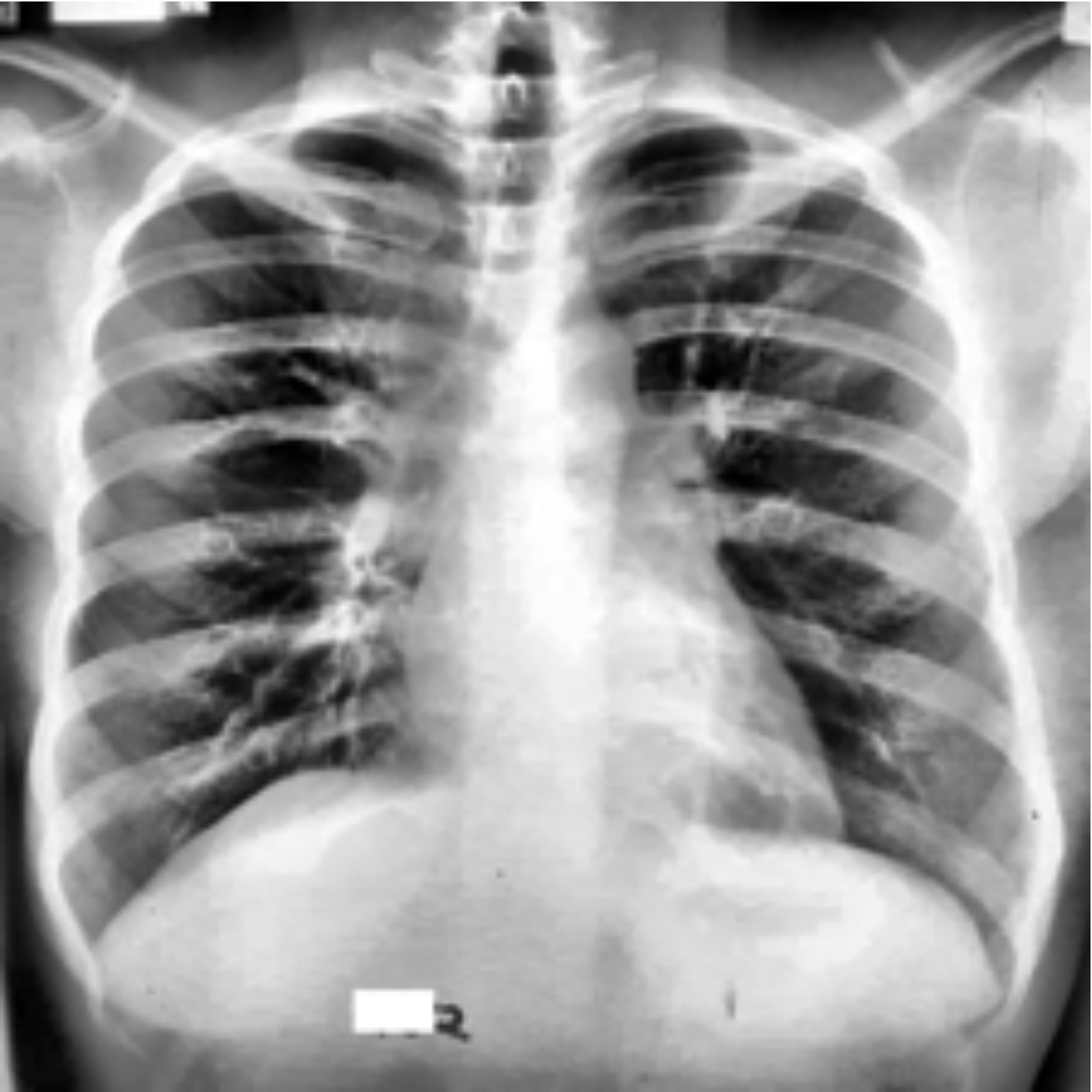}}
\endminipage\hfill
\minipage{0.2\textwidth}
  \subfloat[]{\label{main:a}\includegraphics[scale=.075]{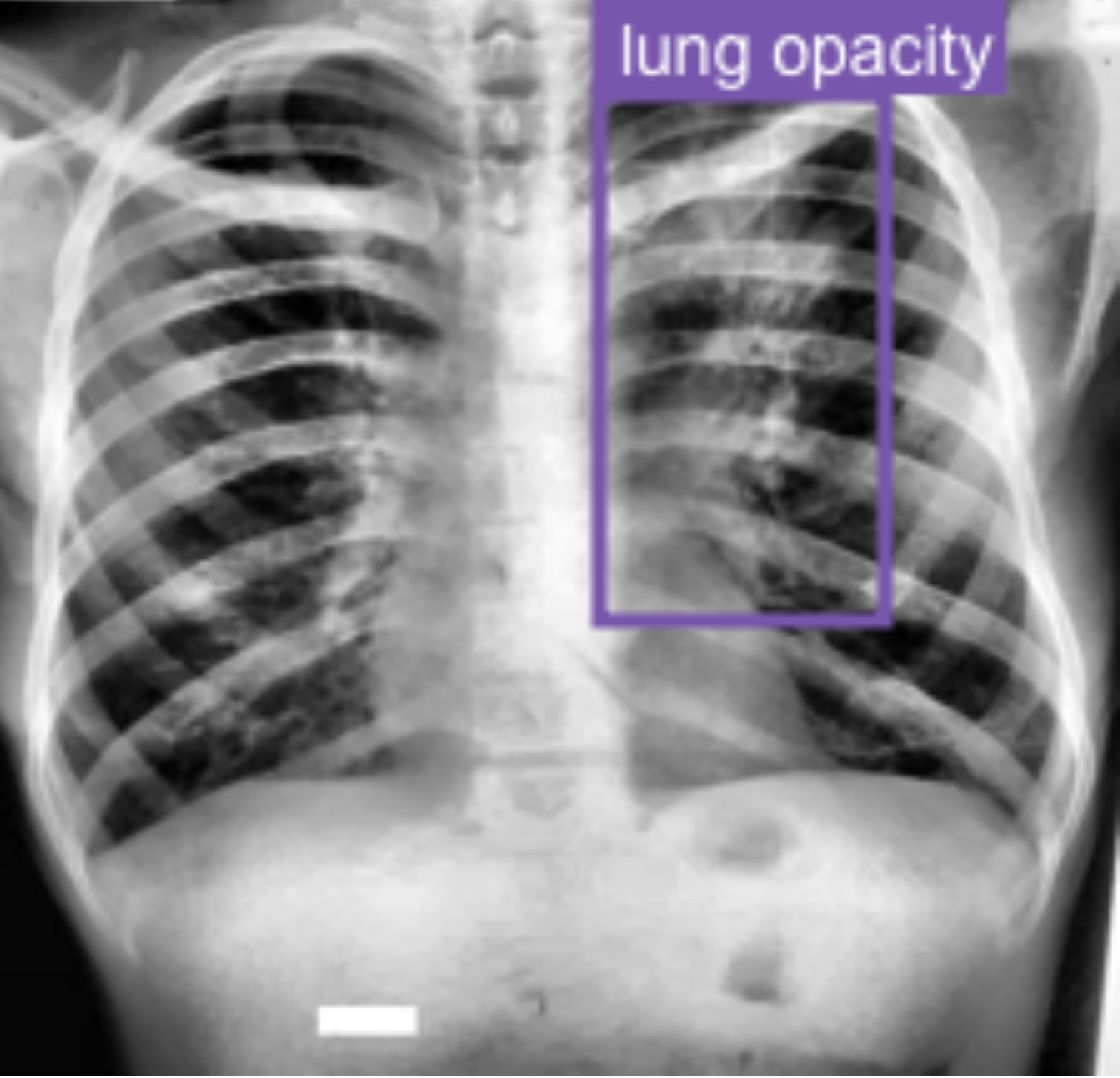}}
\endminipage\hfill
\minipage{0.2\textwidth}
  \subfloat[]{\label{main:a}\includegraphics[scale=.15]{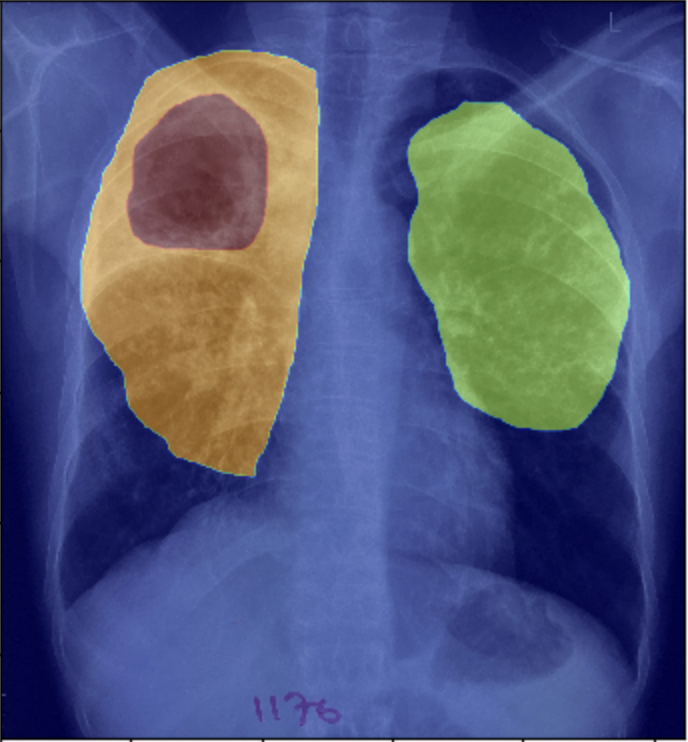}}
\endminipage
\caption{Images from the training dataset. (a) shows an image with active TB, (b) shows a TB negative image, (c) shows an image with latent TB, (d) shows an image with bounding box, and (e) shows an image with segmentation masks.}
\end{figure}

\subsubsection{Disease diagnosis.} We used the TB diagnosis labels to create the disease diagnosis dataset. We structured the prompt as follows, ``<image> What is the diagnosis in the X-ray image?''. The model is expected to directly output the diagnosis of the image.
\subsubsection{Report Generation.}
We created medical reports comprised of findings and impressions sections. The findings section is made up of the pathologies annotated on the image. We used the disease diagnosis as the impression. For images with a TB negative diagnosis, we used ``no findings''. We structured the prompt as follows ``<image> Generate a medical report for the X-ray image provided.'' where <image> is the placeholder for the image.
\subsubsection{Visual Question Answering.}
We created the VQA dataset from the medical reports. We prompted GPT-4o to generate questions and answers in four categories, a subset of those in the VQA-RAD~\cite{Lau2018} dataset that were relevant to our dataset. The categories were abnormality, presence, location, and counting. The dataset contained 623 abnormality, 1,245 presence, 625 position, and 624 counting questions. It included both open-ended and close-ended questions. Close-ended questions required precise answers such as 'yes,' 'no,' or numerical values, while open-ended questions required sentence-based responses. In total, the dataset contained 3,117 questions, comprising 1,902 close-ended and 1,215 open-ended questions. We used these questions as prompts.
\subsubsection{Object Detection.}
We extracted bounding boxes from the segmentation masks in COCO format.
We converted the coordinates into 4 location tokens by dividing them by the image size and multiplying by 1000. The labels were formatted as ``<loc0398><loc0663><loc0773><loc0905> {pathology}''. We structured the prompt as follows: ``<image> detect {pathology}; ...{pathology}''.
\subsubsection{Segmentation.} We used the segmentation masks annotated by the radiologists to create the segmentation dataset. We converted the mask into 16 segmentation tokens using VQ-VAE~\cite{oord2018neuraldiscreterepresentationlearning}. We converted the labels to the following format: ``$<loc0001>...<loc0004><seg001>...<seg016>$ {pathology}'', where the first 4 tokens represent the bounding box and the next 16, the segmentation mask. We structured the prompt as follows: ``segment \{pathology\}..\{pathology\}''. We used the full set of pathologies in the dataset in the prompt.

\section{Method}

\subsection{A Unified Learning Paradigm}
In this section, we describe the learning paradigm for unifying different medical tasks into a generative framework.
In our multimodal dataset, each training sample is composed of two elements: $X = \{T, V\}$, where $T$ refers to the language part with image tokens inserted at the beginning. $V$ refers to 2D images: $V \in R^{H\times W \times C}$, $H, W, C$ are height, width and number of channels respectively, corresponding to the special image tokens in the text $T$. The objective of the model is the likelihood of generated text tokens in $T$, conditioned on the text and images.
\begin{equation}
    p(T|V) = \prod p(T_{i}|V, T_{<i})
\end{equation} 
where $T_{i}$ represents the i-th token in $T$ and $V$ $T_{<i}$ represent the image and language apearing before the i-th token. We used a generative model $\Phi_{LLM}$ to parameterize the probability p, and our final training objective is the negative log-likelihood of the correct next token in the text sequence
$$L_{reg}  = -\sum w_llog\Phi_{LLM}(T_{i}|V T_{<i})$$
where $w_l$ refers to a per-token weighting. $w_l$ is 1 for all response tokens and 0 for prompt and image tokens.
\subsection{Architecture Detail}
We used the PaliGemma~\cite{Beyer2024} architecture. PaliGemma is a foundation vision language model composed of a SigLIP vision encoder, Gemma-2B, a decoder only language model and a linear layer that projects the output of the vision encoder to the same dimension as the Gemma-2B vocab tokens for concatenation. The
text input is tokenized using the Gemma SentencePiece tokenizer and embedded
using Gemma’s embedding layer. The image is encoded using a visual encoder
and projected to the same dimension as the text embeddings. The resulting em-
bedding are concatenated and passed to Gemma for decoding.

\begin{figure}
    \centering
    \includegraphics[width=1.0\linewidth]{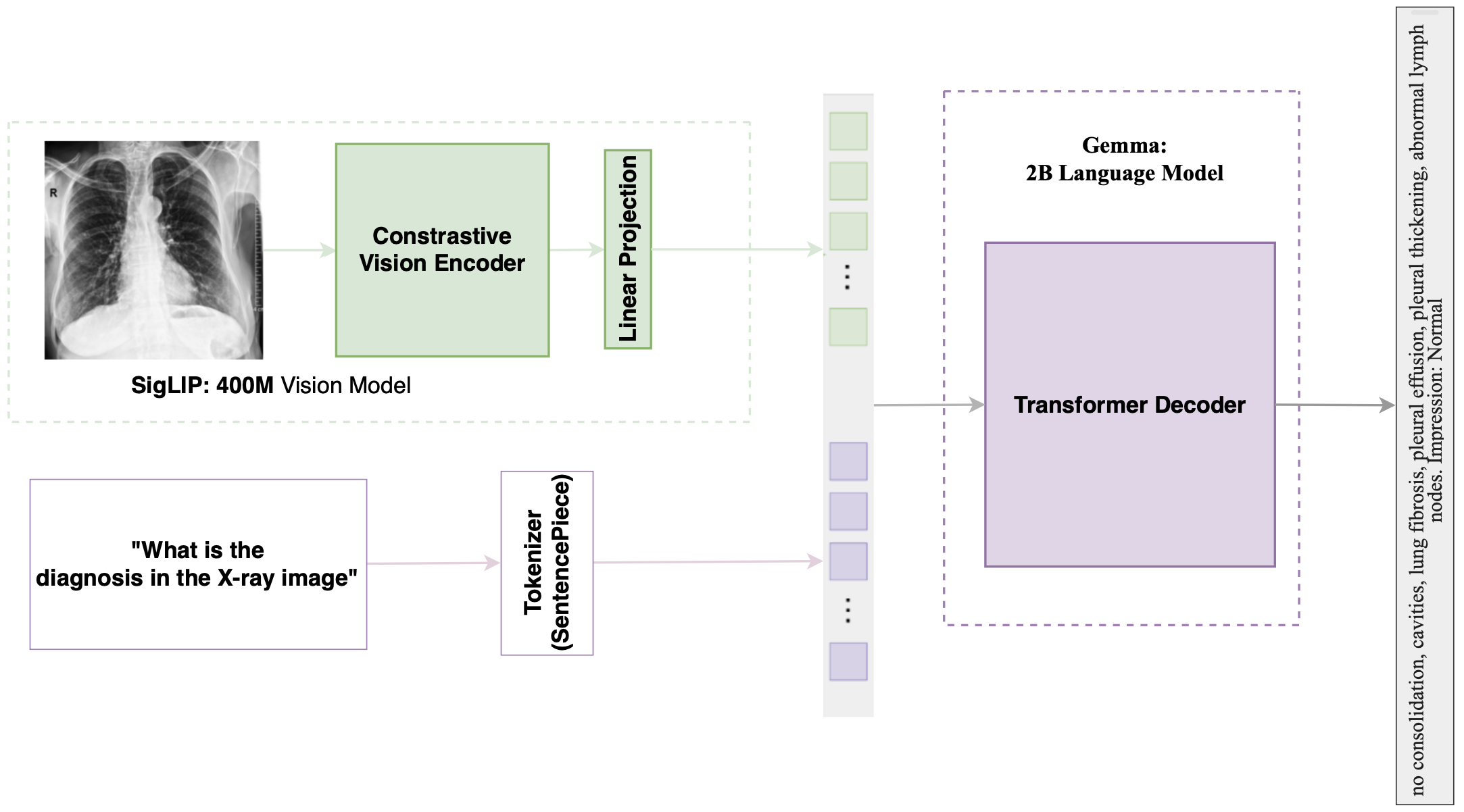}
    \caption{PaliGemma-CXR architecture: SigLIP image encoder feeds into Gemma decoder LM}
    \label{fig:paligemma-arch}
\end{figure}

\subsection{Training Procedure}
To prevent information leakage between the training and testing splits, we randomly split the images into train, validation and test datasets  in ratios of 8:1:1 respectively before creating task specific datasets. This ensures that no images in any of the task training datasets appear in the test dataset of another task.
\subsubsection{Image pre-processing.} We converted all images with a single channel to RGB. We normalized the images using the mean and standard deviation of the images in the training dataset. We resized the images to 224 $\times$ 224 $\times$ 3. 
\subsubsection{Training details}
We finetuned the model jointly on all the five tasks.  Within each epoch, we sampled data from the task specific datasets using a ratio of the inverse of the size of each task-specific dataset. Specifically, the ratios were 6:9:6:1:8 for disease diagnosis, object detection, report generation, VQA and segmentation respectively. We experimented with training the model sequentially on all the datasets but this resulted in catastrophic forgetting. We also experimented with mixing the data within each mini-batch by sampling data in ratios of the inverse of the size of each dataset but this required a large batch size and did not yield any performance improvements over our method.
We initialized the model using the HuggingFace PaliGemma model checkpoint \footnote{https://huggingface.co/google/paligemma-3b-pt-224}. We froze the image encoder and the multimodal encoder and finetuned the language model. In our preliminary experiments, we found that finetuning the other components of the model resulted in sub-optimal performance. 
We finetuned the model using next token prediction with cross entropy loss. We applied the loss to suffix tokens. We used the AdamW optimizer with $(\beta_1 = 0.9, \beta_2 = 0.999)$ and a constant learning rate of 1e-5 and a weight decay of 0.01. We used a batch size of 16 with 4 gradient accumulation steps and trained the model for 10 epochs. We used uniform weights for losses of all tasks. We implemented the model in Pytorch and trained on 1 NVIDIA A40 GPU with 48 GB VRAM. 

\subsection{Evaluation}
\label{metrics}
We used various metrics to evaluate PaliGemma-CXR on different tasks. We used accuracy to evaluate disease diagnosis and close ended visual question answering. In addition to accuracy, we also report macro-recall and macro-precision for these tasks considering class imbalance in our dataset. We evaluated report generation and open ended VQA using BLEU4, METEOR and ROUGE-L. 

We evaluated object detection and segmentation using mean Average Precision (mAP) at IoU > 0.5. Intersection over Union (IoU) measures the area of overlap between the actual and predicted bounding box or mask. 
mAP is defined as the mean of the average precision scores for each class. To calculate the average precision for each class, each bounding box predicted by the model is classified as true positive if IoU > 0.5 and the correct class is predicted, false positive if IoU > 0.5 but the wrong class is predicted and false negative if IoU < 0.5 or no bounding box is predicted or incorrect class is predicted. 

\section{Results}
In this section we present results across the five tasks using the metrics described in Section \ref{metrics}.
We compared the performance of PaliGemma-CXR with vision only models for disease classification and object detection. We also compare PaliGemma-CXR with task-specific PaliGemma and zero-shot PaliGemma.
\subsection{Comparison of PaliGemma-CXR with vision only models for disease diagnosis and object detection}
\subsubsection{Disease Diagnosis.} For disease diagnosis, we compared PaliGemma-CXR with two unimodal baselines, ViT-H and SigLip. SigLip is the visual encoder in the PaliGemma architecture. We trained both unimodal baselines for 50 epochs using the AdamW optimizer with learning rate of 1e-4. PaliGemma-CXR outperformed both baselines on accuracy and macro recall as shown in Table \ref{disease diagnosis-results}. SigLIP obtained the highest macro precision but has the least accuracy. The task-specific PaliGemma model outperformed SigLIP on accuracy but SigLIP obtained a worse macro-recall and macro-precision.

\begin{table}[ht]
    \centering
    \caption{Comparison between PaliGemma-CXR, PaliGemma and other baseline unimodal vision only models. PaliGemma-CXR obtained the highest accuracy and macro-recall. SigLip obtained the highest macro-precision.}
    \begin{tabularx}{\textwidth}{p{3cm}p{3cm}p{2cm}p{2cm}p{2cm}}
    \hline
    Model & PaliGemma-CXR & PaliGemma & ViT-H&SigLip \\
    \hline
    Accuracy& \textbf{90.32\%} & 85.48\%& 82.26\%&77.42\%  \\
    Macro Recall& \textbf{60.78\%}	&59.57\% &54.47\% & 60.41\%   \\
    Macro Precision& 58.56\%	&54.70\% 	& 55.84\%& \textbf{67.78\%}\\
    \hline
    \end{tabularx}
    \label{disease diagnosis-results}
\end{table}

\subsubsection{Object Detection.} For object detection, we compared PaliGemma-CXR to DETA~\cite{ouyangzhang2022nmsstrikes}. We used a DETA model with a ResNet-50 backbone, and trained it for 80 epochs using the AdamW optimizer with a learning rate of 2e-4. PaliGemma-CXR outperformed both the DETA model and the task specific PaliGemma. DETA outperformed the task-specific PaliGemma model by a wide margin and nearly matched the performance of PaliGemma-CXR.
\begin{table}[ht]
\caption{Comparison between PaliGemma-CXR, task-specific PaliGemma and DETA on object detection. PaliGemma-CXR obtained the highest mAP.}
\begin{tabularx}{\textwidth}{p{3cm}p{2cm}p{3cm}p{2cm}p{2cm}}
\hline
Task &Metric & PaliGemma-CXR & Task-specific PaliGemma & DETA \\
\hline
Object detection &mAP & \textbf{20.0} & 7.1& 17.9\\
\hline
\end{tabularx}
\label{samples}
\end{table}

\subsection{Comparison between PaliGemma-CXR, task-specific PaliGemma and zero-shot PaliGemma}
The results showing the comparison between PaliGemma-CXR, task-specific PaliGemma and zero-shot PaliGemma are summarized in Table \ref{eval}. PaliGemma-CXR was trained jointly on all the tasks. Task-specific PaliGemma was trained on each task individually. The zero shot PaliGemma was not finetuned on our dataset.  PaliGemma-CXR outperformed all the other models across all tasks and metrics.

\begin{table}
\centering
\caption{Comparison between PaliGemma-CXR, task-specific PaliGemma, and zero-shot PaliGemma on the five tasks. PaliGemma-CXR outperformed all the models across all metrics.}
\begin{tabularx}{\textwidth}{p{3cm}p{2.5cm}p{2cm}p{2cm}p{2cm}}
\hline
Task &  Metric & PaliGemma-CXR & Task-specific PaliGemma & Zero-shot PaliGemma\\
\hline
 &  Accuracy & \textbf{90.3\%} &85.5\%&67.7\%\\
 Disease diagnosis &  Macro Recall & \textbf{60.8\%} &59.6\%&33.3\% \\
  &  Macro Precision &\textbf{58.6\%}  & 54.7\%&22.6\% \\
\midrule
 &  Accuracy & \textbf{98.9\%} &81.9\%&40.5\%\\
VQA &  Bleu-4 & \textbf{12.0\%} & 0.0\%& 0.0\%\\
 &  Meteor & \textbf{54.9\%} &31.2\%& 30.4\%\\
 &  Rouge-L & \textbf{93.5\%} &62.8\%&57.6\%\\
\midrule
 &  Bleu-4 & \textbf{41.3\%} &32.9\%&0.0\%\\
Report Generation &  Meteor & \textbf{55.3\%} &48.3\%&0.2\%\\
 &  Rouge-L & \textbf{72.2\%} &67.1\%&0.8\%\\
\midrule
Object Detection & mAP &\textbf{19.4}&7.1&0.0\\
\midrule
Segmentation & mAP & \textbf{16.0}&7.2&0.0\\
\hline
\end{tabularx}
\label{eval}
\end{table}


\subsubsection{Visual Question Answering (VQA).} For VQA, PaliGemma-CXR outperformed the task-specific PaliGemma and the zero-shot PaliGemma on both close and open ended questions. The zero-shot PaliGemma model obtained the worst performance. Table \ref{vqa} shows a breakdown of the accuracy of PaliGemma-CXR and task-specific PaliGemma on each category of questions. PaliGemma-CXR outperformed the task-specific PaliGemma across all question categories. There was a significant improvement in performance on position and counting questions. We attribute this improvement in performance to the inclusion of object detection and segmentation tasks which require spatial understanding. 
\begin{table}[ht]
\centering
\caption{Comparison of accuracy between PaliGemma-CXR and PaliGemma on the different categories of visual question answering questions.}
\begin{tabularx}{\textwidth}{p{4cm}p{4cm}p{4cm}}
\hline
Question Type &  PaliGemma-CXR & Task-specific PaliGemma\\
\hline
Close ended &  \textbf{98.96\%} & 81.86\%  \\
Abnormality &  \textbf{97.44\% } & 97.36\%\\
Presence & \textbf{99.19\%} & 89.15\% \\
Position & \textbf{100.0\%}& 23.08\%\\
Counting & \textbf{88.71\%}&46.03\%\\
\hline
\end{tabularx}
\label{vqa}
\end{table}

\subsubsection{Report Generation.}
For report generation, PaliGemma-CXR outperformed both task-specific PaliGemma and zero-shot PaliGemma across all metrics.

\subsubsection{Segmentation.}For segmentation, PaliGemma-CXR outperformed the task-specific PaliGemma model and zero-shot PaliGemma model. Zero-shot PaliGemma model did not produce any valid segmentation masks.

\subsection{Qualitative Results}
In this section, we show the qualitative results for different tasks.\\

\begin{figure}[ht]
    \centering
    \includegraphics[width=1.0\linewidth]{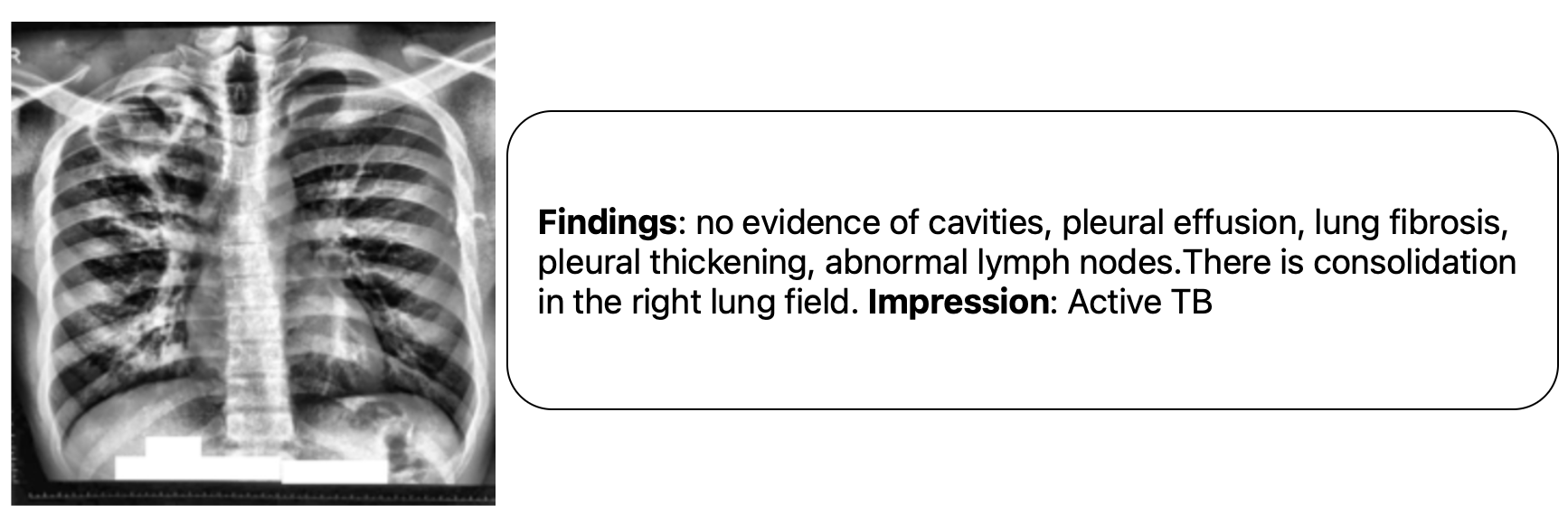}
    \caption{Medical report generated by PaliGemma-CXR. The model identifies consolidation in the right lung field.}
    \label{rrg-image}
\end{figure}

\noindent Figure \ref{rrg-image} shows a medical report generated by PaliGemma-CXR. PaliGemma-CXR identifies the pathologies in the X-ray images and their position.

\begin{figure}[!htb]
    \minipage{0.25\textwidth}
    \centering
    \subfloat[]{\label{main:a}\includegraphics[scale=.16]{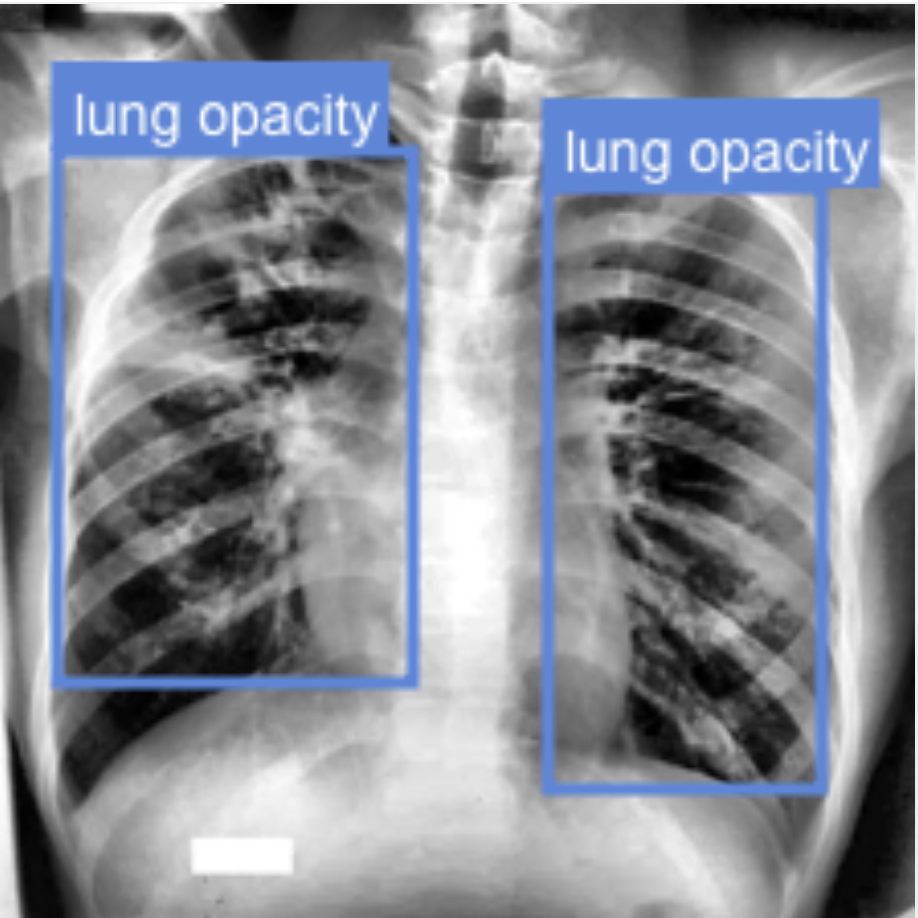}}
    \endminipage
    \minipage{0.25\textwidth}
    \centering
    \subfloat[]{\label{main:a}\includegraphics[scale=.16]{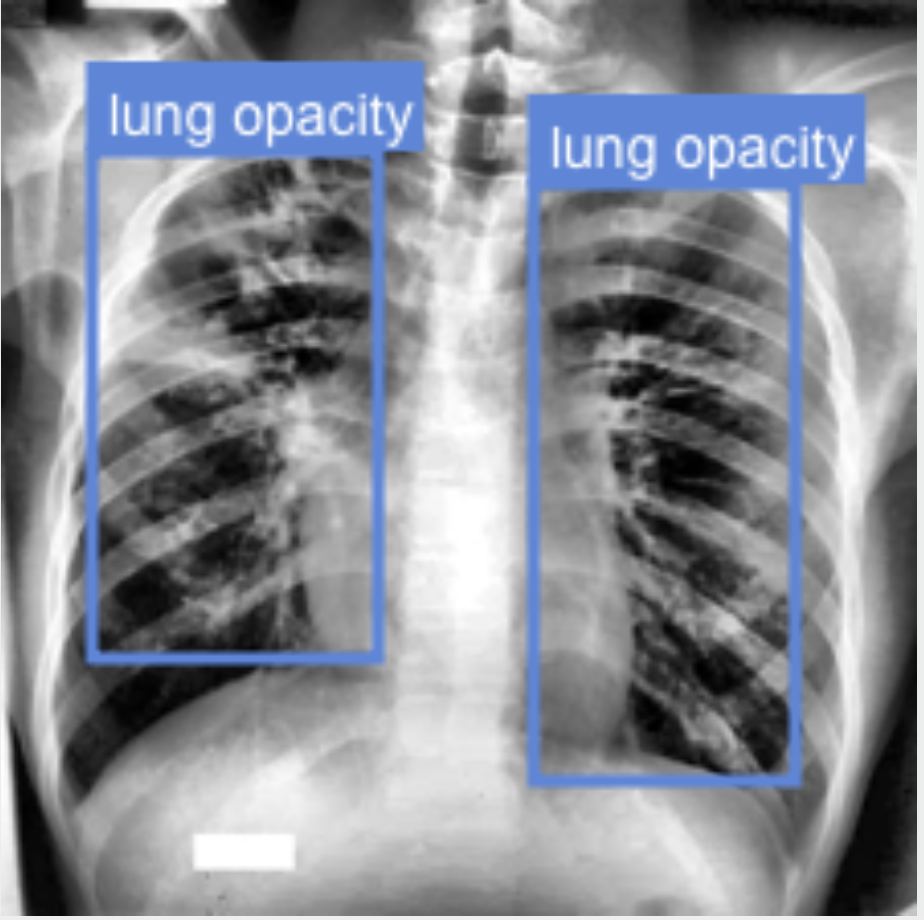}}
    \endminipage
    \label{od-results}
    \minipage{0.25\linewidth}
    \centering
    \subfloat[]{\label{main:a}\includegraphics[scale=.1]{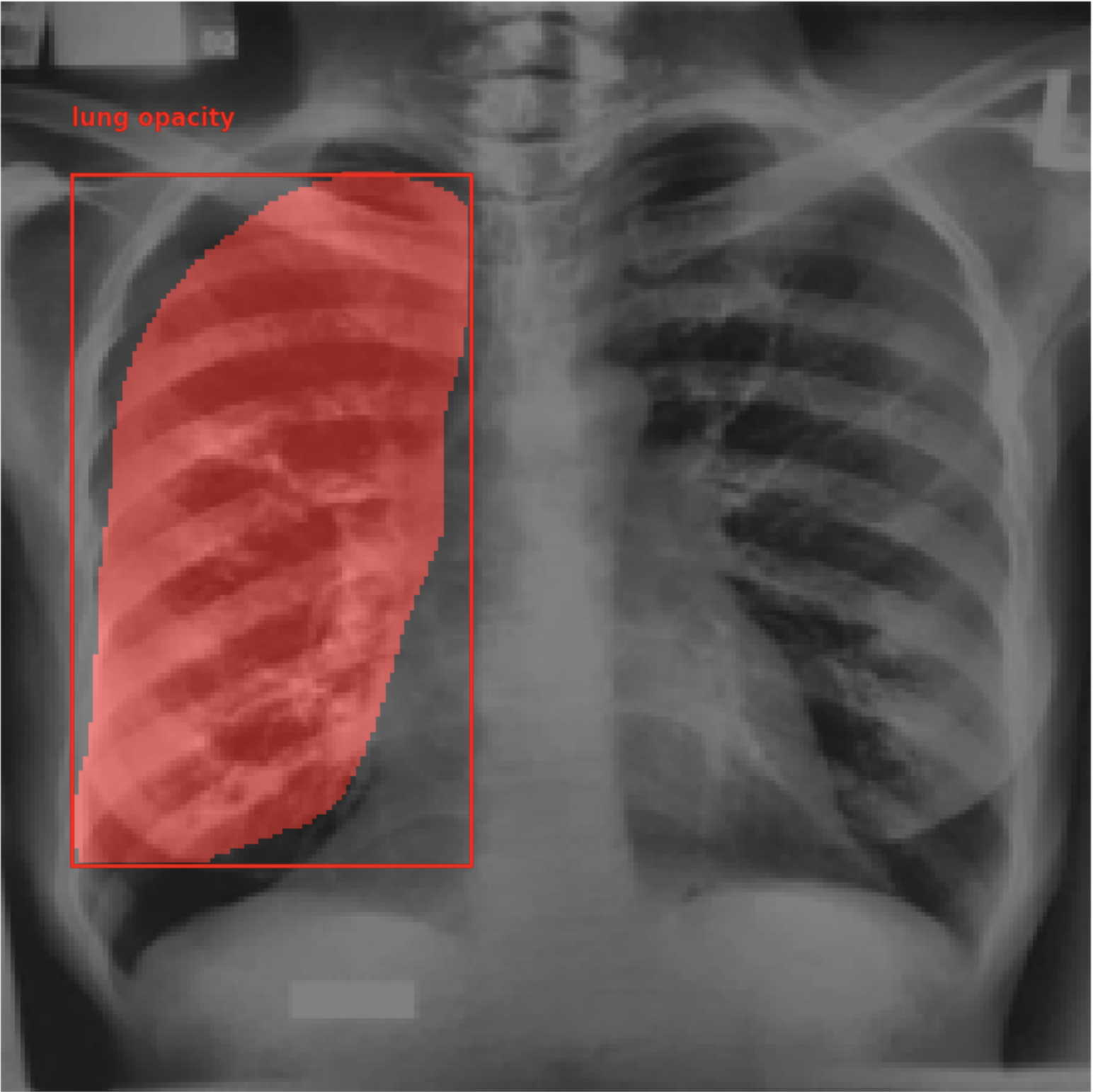}}
    \endminipage
    \minipage{0.25\linewidth}
    \centering
    \subfloat[]{\label{main:a}\includegraphics[scale=.1]{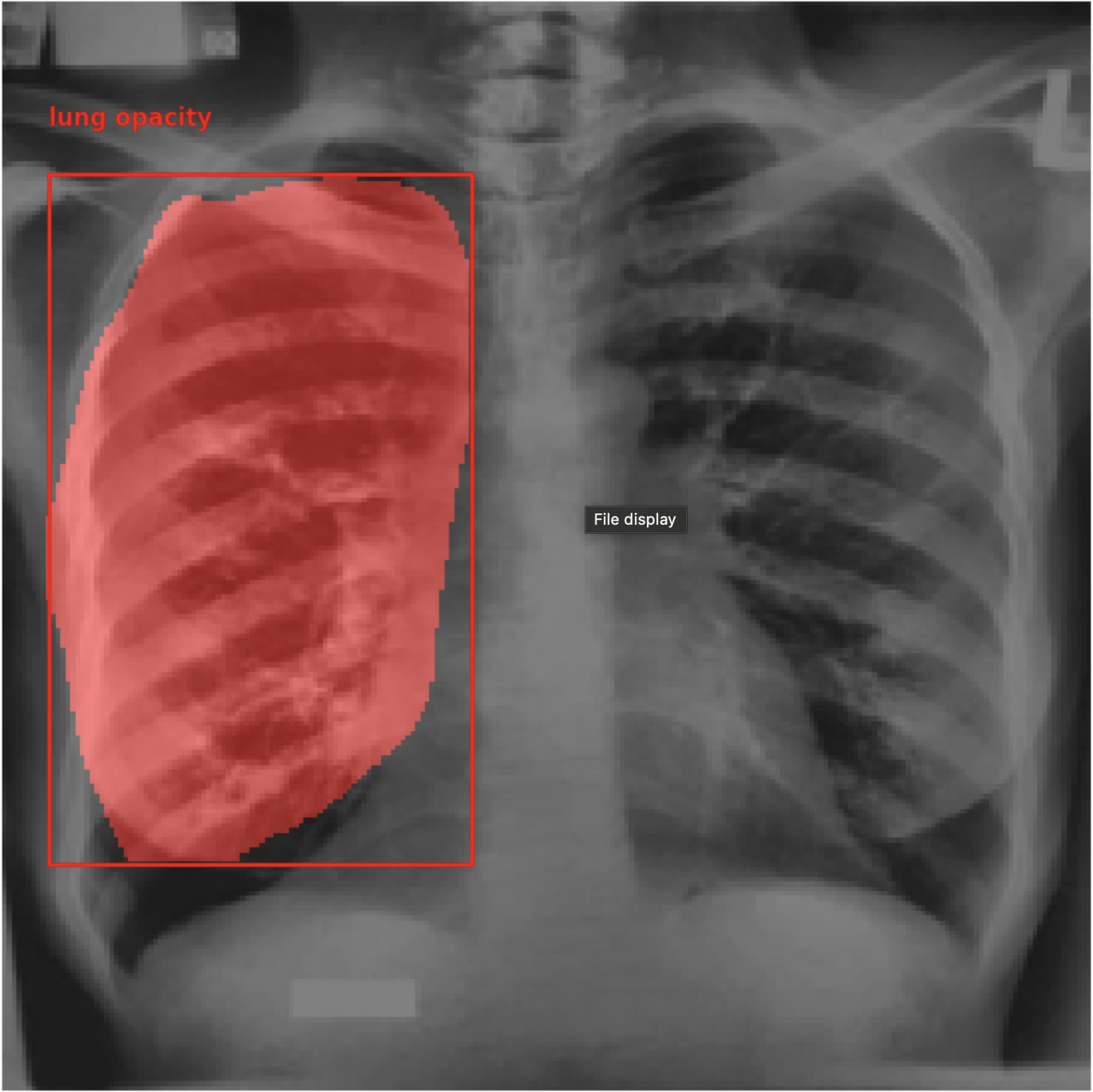}}
    \endminipage
    \caption{(a) shows the bounding boxes predicted by PaliGemma-CXR and (b) shows the ground truth bounding boxes, (c) shows the ground truth segmentation mask and (d) shows the segmentation mask generated by PaliGemma-CXR.}
    \label{od-and-seg}
 \end{figure}

\noindent Figure \ref{od-and-seg} shows the bounding boxes predicted by PaliGemma-CXR. PaliGemma detects the pathologies in the image. It can detect multiple instances of the same pathology as well as different pathologies occurring within the same image.

\noindent Figure \ref{od-and-seg}, shows the segmentation masks predicted by PaliGemma-CXR. It successfully segments predicts segmentation masks with a high IoU with the ground truth segmentation mask.
 
\section{Conclusion}
In this work, we develop a multi-task multimodal model for the interpretation of TB chest X-ray images. We train the model by jointly fine-tuning a base PaliGemma model on five medical image analysis tasks—including classification, report generation, visual question answering, object detection, and segmentation.

To address task imbalance during training, we explored multiple sampling strategies. Our final approach involved sampling data from each task in separate batches, with sampling probabilities set as the inverse of each dataset’s size. This strategy effectively oversampled tasks with smaller datasets, ensuring balanced representation during training. In contrast, a sequential training approach—where we trained on one dataset after another—resulted in catastrophic forgetting, and mixing data within each mini-batch required very large batch sizes without yielding performance improvements. Overall, the inverse dataset-size sampling strategy significantly enhanced model performance, underscoring the importance of balanced task representation.

Across all tasks, PaliGemma-CXR outperformed task-specific PaliGemma and zero-shot PaliGemma. This demonstrates positive transfer between the five tasks when jointly trained together, with performance improvements most notable for object detection and segmentation (gains of 177\% and 122\%, respectively).

PaliGemma-CXR further illustrates the effectiveness of multi-task learning in low-data settings, a critical advantage in medical imaging where labeled data is often scarce. By leveraging shared representations across tasks, our approach outperforms models trained on individual tasks. Notably, our findings contrast with those of Standley et al.~\cite{Standley2019}, who observed that multi-task training degraded performance when using only 5\% of the available data.

A key practical advantage of PaliGemma-CXR is its ability to perform tasks using only a text prompt, eliminating the need for additional inputs. In contrast, models like Med-SAM~\cite{Ma2024} and SAM~\cite{Kirillov2023} —while highly effective for segmentation—require bounding boxes or coordinates to define regions of interest, increasing both effort and deployment time in clinical settings. PaliGemma-CXR’s prompt-based approach simplifies this process, making it more practical and efficient for real-world applications.

\newpage

\bibliographystyle{splncs04}

\end{document}